\documentclass[conference]{IEEEtran}
\IEEEoverridecommandlockouts
\usepackage{cite}
\usepackage{amsmath,amssymb,amsfonts}
\usepackage{algorithm,algorithmic}
\usepackage{graphicx}
\usepackage{textcomp}
\usepackage{xcolor}
\usepackage{footnote}
\usepackage{hyperref}
\usepackage[subtle,tracking=normal]{savetrees}

\def\BibTeX{{\rm B\kern-.05em{\sc i\kern-.025em b}\kern-.08em
		T\kern-.1667em\lower.7ex\hbox{E}\kern-.125emX}}
\begin{document}
	
	\title{Conditional Measurement Density Estimation in Sequential Monte Carlo via Normalizing Flow\\
	}
	
	\author{\IEEEauthorblockN{Xiongjie Chen}
		\IEEEauthorblockA{\textit{Department of Computer Science} \\
			\textit{University of Surrey}\\
			Guildford, UK \\
			xiongjie.chen@surrey.ac.uk}
		\and
		\IEEEauthorblockN{Yunpeng Li}
		\IEEEauthorblockA{\textit{Department of Computer Science} \\
			\textit{University of Surrey}\\
			Guildford, UK \\
			yunpeng.li@surrey.ac.uk}
	}
	
	\maketitle
	
	\begin{abstract}
		Tuning of measurement models is challenging in real-world applications of
		sequential Monte Carlo methods. Recent advances in differentiable 
		particle filters have led to various efforts to learn measurement models 
		through neural networks. But existing approaches in the 
		differentiable particle filter framework do not admit valid probability 
		densities in constructing measurement models, leading to incorrect 
		quantification of the measurement uncertainty given state information. 
		We propose to learn expressive and valid probability densities in 
		measurement models through conditional normalizing flows, to capture 
		the complex likelihood of measurements given states. We show that the 
		proposed approach leads to improved estimation performance and faster training convergence in a visual tracking experiment. 
	\end{abstract}
	
	
	\section{Introduction}
	\label{sec:introduction}
	In this section we first provide a brief introduction to the conditional density estimation (CDE) problem and sequential Monte Carlo (SMC) methods. We then illustrate the role of conditional measurement density estimation in SMC methods, in particular differentiable particle filters (DPFs), and discuss their limitations. We summarize our contributions and outline the structure of this paper at the end of this section.
	
	\subsection{Conditional Density Estimation}
	Conditional density estimation (CDE) refers to the problem of modeling the conditional probability density $p(\textbf{y}|\textbf{x})$ for a dependent variable \textbf{y} given a conditional variable \textbf{x}. In contrast to tasks targeting the conditional mean $\mathbb{E}_{p(\textbf{y}|\textbf{x})}[\textbf{y}]$, CDE aims to model the full conditional density $p(\textbf{y}|\textbf{x})$. Applications of CDE can be found in various domains including computer vision~\cite{sohn2015learning}, econometrics~\cite{malik2011particle}, and reinforcement learning~\cite{bellemare2017distributional}.

	Most CDE approaches can be classified into two categories, parametric CDE models and non-parametric CDE models. 
	With parametric models, an important assumption is that the conditional probability density $p(\textbf{y}|\textbf{x})$ belongs to a family of probability distributions $\mathcal{F}:=\{p(\cdot|\textbf{x};\theta)|\theta\in\Theta\}$, where the distribution $p(\cdot|\textbf{x};\theta)$ is uniquely determined by a parameter set $\theta$ and a conditional variable $\textbf{x}$. In particular, $p(\textbf{y}|\textbf{x};\theta)$ is often formulated as a parameterized function $l_\theta(\textbf{y},\, \textbf{x})$ whose parameters $\theta$ are learned from data by minimizing the negative log likelihood (NLL) of training data. Examples of parametric CDE models include the conditional normalizing flow (CNF)~\cite{lu2020structured,winkler2019learning}, the conditional variational autoencoder (CVAE)~\cite{sohn2015learning}, the mixture density network (MDN)~\cite{bishop1994mixture}, and the bottleneck conditional density estimator (BCDE)~\cite{shu2017bottleneck}. In contrast, non-parametric CDE models do not impose any parametric restrictions on the distribution family that $p(\textbf{y}|\textbf{x})$ belongs to. So, theoretically they can approximate arbitrarily complex conditional densities~\cite{ambrogioni2017kernel}. Specifically, non-parametric CDE approaches often describe the distribution $p(\textbf{y}|\textbf{x})$ by specifying a kernel, e.g. the Gaussian kernel, to each training data sample. However, most non-parametric CDE models assume $p(\textbf{y}|\textbf{x})$ to be smooth, and require traversing the entire training set to make a single prediction~\cite{trippe2018conditional}. Examples of non-parametric CDE models include the kernel mixture network (KMN)~\cite{ambrogioni2017kernel}, the conditional kernel density estimation (CKDE)~\cite{bashtannyk2001bandwidth}, and the Gaussian process conditional density estimation~\cite{dutordoir2018gaussian}.
	
	\subsection{Likelihood Learning in Sequential Monte Carlo methods}
	While CDE models have been employed in many practical applications, in this work we focus on CDE models to describe the relation between hidden states and observations in sequential Monte Carlo methods, i.e. measurement models in particle filters.
	
	We first introduce the problem setup. Sequential Monte Carlo methods, a.k.a. particle filters, are a set of powerful and flexible simulation-based methods designed to numerically solve sequential state estimation problems~\cite{gordon1993novel}. The sequential state estimation problem we consider here is characterized by an unobserved state $\{\textbf{x}_t\}_{t\geq 0}$ defined on $\mathcal{X}\subseteq\mathbb{R}^{d_\textbf{x}}$ and an observation $\{\textbf{y}_t\}_{t\geq 1}$ defined on $\mathcal{Y}\subseteq\mathbb{R}^{d_\textbf{y}}$. The evolution of $\textbf{x}_t$ and the relationship between $\textbf{y}_t$ and $\textbf{x}_t$ can be described by the following Markov process:
	\begin{align}
		\textbf{x}_0&\sim \pi(\textbf{x}_0)\,\,,\\
		\textbf{x}_t&\sim p(\textbf{x}_t| \textbf{x}_{t-1},\textbf{a}_t;\theta) \text{ for } t\geq1\,\,,\label{eq:transition}\\
		\textbf{y}_t&\sim p(\textbf{y}_t| \textbf{x}_t;\theta) \text{ for } t\geq1\,\,, \label{eq:likelihood}
	\end{align}
	where $\theta\in\Theta$ denotes the parameter set of interest, $\pi(\cdot)$ is the initial distribution, $p(\textbf{x}_t| \textbf{x}_{t-1},\textbf{a}_t;\theta)$ is the dynamic model, $\textbf{a}_t$ is the action, $p(\textbf{y}_t| \textbf{x}_t;\theta)$ is the measurement model estimating the likelihood~\footnote{For brevity, we use the term ``likelihood'' and ``conditional measurement density'' interchangeably for the rest of the paper.} of $\textbf{y}_t$ given $\textbf{x}_t$. Denote by $\textbf{x}_{0:t}\triangleq\{\textbf{x}_0,\,\,\cdots,\,\,\textbf{x}_t\}$, $\textbf{y}_{1:t}\triangleq\{\textbf{y}_1,\,\,\cdots,\,\,\textbf{y}_t\}$ and $\textbf{a}_{1:t}\triangleq\{\textbf{a}_1,\,\,\cdots,\,\,\textbf{a}_t\}$ sequences of hidden states, observations and actions up to time step $t$, respectively. Our goal is to simultaneously track the joint posterior distribution $p(\textbf{x}_{0:t}|\textbf{y}_{1:t}, \textbf{a}_{1:t}; \theta)$ or the marginal posterior distribution $p(\textbf{x}_{t}|\textbf{y}_{1:t}, \textbf{a}_{1:t}; \theta)$ and learn the parameter $\theta$ from data. However, except in some trivial cases such as linear Gaussian models, the posterior is usually analytically intractable.
	
	Particle filters approximate the analytically intractable posteriors with empirical distributions consisting of sets of weighted samples:
	\begin{equation}
		{p}(\textbf{x}_{0:t}|\textbf{y}_{1:t}, \textbf{a}_{1:t}; \theta)\approx \sum_{i=1}^{N_p}w_t^i \delta_{\textbf{x}_{0:t}^i}(d\textbf{x}_{0:t})\,\,,
	\end{equation}
	where $N_p$ is the number of particles, $\delta_{\textbf{x}_{0:t}^i}(\cdot)$ denotes the Dirac measure located in $\textbf{x}_{0:t}^i$, $w_t^i$ is the importance weight of the $i$-th particle at time step $t$.	Sampled from the initial distribution $\pi(\textbf{x}_0)$ at time step $t=0$ and proposal distributions $q(\textbf{x}_{t}|\textbf{x}_{t-1}, \textbf{y}_{t}, \textbf{a}_{t};\theta)$ at time step $t\geq 1$, the importance weight $w_t^i$ of the $i$-th particle $\textbf{x}_{0:t}^i$ can be computed recursively for each time step:
	\begin{align}\label{eq:recursive_weight}
		w_t^i\propto w_{t-1}^i \frac{p(\textbf{y}_{t}|\textbf{x}_{t}^i;\theta) p(\textbf{x}_t^i|\textbf{x}_{t-1}^i, \textbf{a}_t; \theta)}{q(\textbf{x}_t^i|\textbf{x}_{t-1}^i, \textbf{y}_{t}, \textbf{a}_{t};\theta)}\,\,, \text{ for } t\geq1\,\,,
	\end{align}
	with $w_0^i=\frac{1}{N_p}$.
	The evaluation of the likelihood $p(\textbf{y}_t|\textbf{x}_t^i; \theta)$ is required at every time step in SMC methods as shown in Eq.~\eqref{eq:recursive_weight}. Traditionally, the measurement model needs to be handcrafted. However, when observations are high-dimensional and unstructured, e.g. images, the specification of measurement models for SMC methods can be challenging and often relies on extensive domain knowledge. Estimation of model parameters in particle filters is an active research area~\cite{malik2011particle, doucet2003parameter, andrieu2005line}. Many existing approaches are restricted to models where the model structure or a subset of model parameters is known~\cite{kantas2015particle}.

	Differentiable particle filters (DPFs) were proposed to alleviate the challenges in handcrafting the dynamic models and measurement models in SMC methods~\cite{karkus2018particle,jonschkowski18,corenflos2021differentiable, wen2021end, chen2021differentiable}. Particularly, the measurement model in DPFs is learned from data by combining the algorithmic priors in SMC methods with the expressiveness of neural networks. For example,~\cite{jonschkowski18} considers the observation likelihoods as the scalar outputs of a fully-connected neural network whose input is the concatenation of states and encoded features of observations. In~\cite{karkus2018particle}, feature maps of both observations and states are fed into a fully connected neural network whose outputs are used as the likelihood of the observations. The likelihood of observations in~\cite{wen2021end,chen2021differentiable} is constructed as the cosine similarity between the feature maps of observations and states. In~\cite{corenflos2021differentiable}, the likelihood is given by a multivariate Gaussian model outputting the Gaussian density of observation feature vectors conditioned on state feature vectors. In complex environments such as robot localization tasks with image observations, DPFs have shown promising results~\cite{karkus2018particle,jonschkowski18,corenflos2021differentiable, wen2021end, chen2021differentiable}. Nonetheless, to the best of our knowledge, existing DPFs do not admit valid probability densities in likelihood estimation, leading to incorrect 
	quantification of the measurement uncertainty given state information.
	
	
	In this paper, we present a novel approach to learn the conditional measurement density in DPFs. Our contributions are three-fold: 1)  We developed a novel approach to construct measurement models capable of modeling complex likelihood with valid probability densities through conditional normalizing flows; 2) We show how to incorporate and train the conditional normalizing flow-based measurement model into existing DPF pipelines; 3) We show that the proposed method can improve the performance of state-of-the-art DPFs in a visual tracking experiment.
	
	The remainder of the paper is organized as follows: Section \ref{sec:background} provides background knowledge and Section \ref{sec:proposed_method} describes the proposed conditional normalizing flow-based measurement model. We report the simulation setup
	and experiment results in Section \ref{sec:experiment_results} and conclude the paper in Section \ref{sec:conclusion}.
	
	\section{Background}
	\label{sec:background}

	\subsection{Differentiable Particle Filters}
	\label{subsec:dpf}
	In differentiable particle filters, both the dynamic model and the measurement model are often formulated as parameterized models. For example, the transition of states and the likelihood of observations can be modeled by the following parameterized functions, respectively:
	\begin{align}
		\label{eq:dpf_dynamic_model}
		\textbf{x}_t^i&=g_\theta(\textbf{x}_{t-1}^i, \textbf{a}_t, \alpha_t^i)\sim p(\textbf{x}_t|\textbf{x}_{t-1}^i,\textbf{a}_t; \theta)\,\,,\\
		\label{eq:dpf_measurement_model}
		\textbf{y}_t& \sim p(\textbf{y}_t|\textbf{x}_t^i; \theta)=l_\theta(\textbf{y}_t, \textbf{x}_t^i)\,\,,
	\end{align}
	where $\alpha_t^i$ is the process noise. In~\cite{jonschkowski18,karkus2018particle,wen2021end,corenflos2021differentiable, chen2021differentiable}, $g_\theta(\cdot)$ is parameterized by neural networks taking $\textbf{x}_{t-1}$ and $\textbf{a}_t$ as inputs.~\cite{corenflos2021differentiable} adopted different dynamic models in different experiments, including multivariate Gaussian models whose mean and covariance matrices are functions of $\textbf{x}_{t-1}$ and $\textbf{a}_t$, and a pre-defined physical model designed based on prior knowledge of the system. For the measurement model, $l_\theta(\textbf{y}_t, \textbf{x}_t^i)$ is designed as neural networks with scalar outputs in~\cite{jonschkowski18,karkus2018particle,wen2021end,chen2021differentiable}, and multivariate Gaussian models in~\cite{corenflos2021differentiable}.
	In addition, proposal distributions can also be specified through a parameterized function $f_\theta(\cdot)$:
	\begin{equation}
		\label{eq:dynamics_semi}
		\textbf{x}_t^i=f_\theta(\textbf{x}_{t-1}^i, \textbf{y}_{t}, \textbf{a}_t, \beta_t^i)\sim q(\textbf{x}_t^i|\textbf{x}_{t-1}^i, \textbf{y}_{t}, \textbf{a}_{t};\theta)\,\,,
	\end{equation}
	where $\beta_t^i$ is the noise term used to generate proposed particles. In~\cite{chen2021differentiable}, $f_\theta(\cdot)$ is built with conditional normalizing flows to move particles to regions close to posterior distributions.
	
	With the particle filtering framework being parameterized, DPFs then optimize their parameters by minimizing certain loss functions via gradient descent. Different optimization objectives have been proposed for training DPFs, which can be categorized as supervised loss~\cite{jonschkowski18,karkus2018particle} and semi-supervised loss~\cite{wen2021end}. Supervised losses such as the root mean square error (RMSE) and the negative log-likelihood (NLL) require access to the ground truth states, and then compute the difference between the predicted state and the ground truth~\cite{jonschkowski18,karkus2018particle}. In~\cite{wen2021end}, a pseudo-likelihood loss was proposed to enable semi-supervised learning for DPFs, which aims to improve the performance of DPFs when only a small portion of data are labeled.
	
	\subsection{Conditional Density Estimation via Conditional Normalizing Flows}
	
	Normalizing flows are a family of invertible mappings transforming probability distributions. A simple base distribution can be transformed into arbitrarily complex distribution using normalizing flows under mild assumptions~\cite{papamakarios2021normalizing}. Since the transformation is invertible, it is straightforward to evaluate the density of the transformed distribution by applying the change of variable formula.
	
	One application in which normalizing flows is particularly well suited is density estimation~\cite{dinh2016real,papamakarios2021normalizing,winkler2019learning}. Denote by $p(\textbf{y})$ a probability distribution defined on $\mathcal{Y}\subseteq\mathbb{R}^{d_\textbf{y}}$ and $\mathcal{T}(\cdot):\mathbb{R}^{d_\textbf{y}}\rightarrow\mathbb{R}^{d_\textbf{y}}$ a normalizing flow. By applying the normalizing flow $\mathcal{T}(\textbf{y})$, the probability density $p(\textbf{y})$ can be evaluated as:
	\begin{equation}
		\label{eq:density_estimation}
		p(\textbf{y})=p_Z\big(\mathcal{T}(\textbf{y})\big)\bigg|\text{det}\frac{\partial \mathcal{T}(\textbf{y})}{\partial \textbf{y}}\bigg|\,\,,	
	\end{equation}
	where $p_Z(\cdot)$ is a base distribution~\cite{papamakarios2021normalizing}, and $\big|\text{det}\frac{\partial \mathcal{T}(\textbf{y})}{\partial \textbf{y}}\big|$ is the absolute value of the determinant of the Jacobian matrix $\frac{\partial \mathcal{T}(\textbf{y})}{\partial \textbf{y}}$ evaluated at $\textbf{y}$.
	As a special case of density estimation, conditional density estimation (CDE) can be realized by using variants of normalizing flows, e.g. conditional normalizing flows (CNFs)~\cite{winkler2019learning,chen2021differentiable,lu2020structured}. Similar to Eq.~\eqref{eq:density_estimation} and with a slight abuse of notations, a conditional normalizing flow $\mathcal{T}(\textbf{y},\textbf{x}):\mathbb{R}^{d_\textbf{y}}\times \mathbb{R}^{d_\textbf{x}}\rightarrow\mathbb{R}^{d_\textbf{y}}$ can be constructed to evaluate the conditional probability density $p(\textbf{y}|\textbf{x})$:
	\begin{equation}
		\label{eq:conditional_density_estimation}
		p(\textbf{y}|\textbf{x})=p_Z\big(\mathcal{T}(\textbf{y},\textbf{x})\big)\bigg|\text{det}\frac{\partial \mathcal{T}(\textbf{y}, \textbf{x})}{\partial \textbf{y}}\bigg|\,\,,
	\end{equation}
	where $\textbf{x}$ is the variable on which $p(\textbf{y}|\textbf{x})$ is conditioned.
	
	Compared with non-parametric and other parametric CDE models, CNFs are efficient and can model arbitrarily complex conditional densities. They have been employed to estimate complex and high-dimensional conditional probability densities~\cite{lu2020structured,winkler2019learning,trippe2018conditional,abdelhamed2019noise}. Therefore, we choose to use CNFs to construct the measurement model in DPFs.
	
	\section{Differentiable Particle Filter with Conditional Normalizing Flow Measurement model}
	\label{sec:proposed_method}
	
	We provide in this section the details of the proposed method and a generic DPF framework where our measurement model is incorporated. The proposed measurement model is built based on conditional normalizing flows, which can model expressive and valid probability densities and capture the complex likelihood of observations given state information.
	
	\subsection{Likelihood Estimation with Conditional Normalizing Flows}
	Consider an SMC model where we have an observation $\textbf{y}_t$ and a set of particles $\{\textbf{x}_t^i\}_{i=1}^{N_p}$ at the $t$-th time step. The proposed measurement model estimates the likelihood of $\textbf{y}_t$ given $\textbf{x}_t^i$ by:
 	\begin{align}
	 	\label{eq:conditional_density_estimation_obs}
	 	p(\textbf{y}_t|\textbf{x}_t^i;\theta)=p_Z(\mathcal{T}_\theta(\textbf{y}_t, \textbf{x}_t^i))\bigg|\text{det}\frac{\partial \mathcal{T}_\theta(\textbf{y}_t, \textbf{x}_t^i)}{\partial \textbf{y}_t}\bigg|\,\,,
	 \end{align}
    where $\mathcal{T}_\theta(\textbf{y}_t,\textbf{x}_t^i)$ is a parameterized conditional normalizing flow. The base distribution $p_Z(\cdot)$ can be user-specified and is often chosen as a simple distribution such as isotropic Gaussian.
    
    In scenarios where the observations are high-dimensional such as images, evaluating Eq.~\eqref{eq:conditional_density_estimation_obs} with the raw observations $\textbf{y}_t$ can be computationally expensive. As an alternative solution, we can map the observation $\textbf{y}_t$ to a lower-dimensional space via $\textbf{e}_t=E_\theta(\textbf{y}_t)\in\mathbb{R}^{d_\textbf{e}}$, where $E_\theta$ is a parameterized function $E_\theta(\cdot):\mathbb{R}^{d_\textbf{y}}\rightarrow\mathbb{R}^{d_\textbf{e}}$. Assume that the conditional probability density $p(\textbf{e}_t|\textbf{x}^i_t;\theta)$ of observation features given state is an approximation of the actual measurement likelihood $p(\textbf{y}_t|\textbf{x}_t^i;\theta)$, we then estimate $p(\textbf{y}_t|\textbf{x}_t^i;\theta)$ with $p(\textbf{e}_t|\textbf{x}^i_t;\theta)$:
 	\begin{align}
	 	\label{eq:conditional_density_estimation_features}
	 	p(\textbf{y}_t|\textbf{x}_t^i;\theta)&\approx p(\textbf{e}_t|\textbf{x}^i_t;\theta)\\
	 	&=p_Z(\mathcal{T}_\theta(\textbf{e}_t, \textbf{x}_t^i))\bigg|\text{det}\frac{\partial \mathcal{T}_\theta(\textbf{e}_t, \textbf{x}_t^i)}{\partial \textbf{e}_t}\bigg|\,\,.
	 \end{align}
 	
 	\subsection{Numerical Implementation}
	\begin{algorithm}[t]
 		\begin{algorithmic}[1]
 			\caption{A generic differentiable particle filter with conditional normalizing flow measurement model (DPF-CM) framework. $[N_p]$ is used to denote the set \{1, 2, $\cdots$ , $N_p$\}.}\label{alg:DPFs-CM}
 			\STATE Initialize the parameter set $\theta$ and set learning rate $\gamma$;
 			\WHILE{$\theta$ has not converged}
 			\STATE Draw samples $\{\textbf{\textbf{x}}_0^i\}_{i=1}^{N_p}\sim \pi(\textbf{x}_0)$;
 			\STATE Set importance weights $w_0^i=\frac{1}{N_p}$ for $i\in[N_p]$;	
 			\STATE Set $\tilde{\textbf{x}}_0^i=\textbf{x}_0^i$ for $i\in[N_p]$;
 			
 			\FOR {$t=1, 2, \dots ,T$}
 			\STATE Sample $\textbf{x}_t^i$ from proposal distribution for $i\in[N_p]$:\\
 			$\textbf{x}_t^i=f_\theta(\tilde{\textbf{x}}_{t-1}^i, \textbf{y}_{t}, \textbf{a}_t, \beta_t^i)\sim q(\textbf{x}^i_t|\tilde{\textbf{x}}_{t-1}^i, \textbf{y}_t, \textbf{a}_t;\theta)$;
 			\STATE [optional:] Encode observation:
 			$\textbf{y}_t :=E_\theta(\textbf{y}_t)$;
 			\STATE Apply conditional normalizing flows: $\textbf{z}_t^i=\mathcal{T}_\theta(\textbf{y}_t,\textbf{x}_t^i)$;
 			\STATE Estimate observation likelihood for $i\in[N_p]$:\\ $p(\textbf{y}_t|\textbf{x}_t^i;\theta)=p_Z(\textbf{z}_t^i)\bigg|\text{det}\frac{\partial \textbf{z}_t^i}{\partial \textbf{y}_t}\bigg|$;
 			\STATE Calculate importance weights for $i\in[N_p]$:\\ $w_t^i=w_{t-1}^i \frac{p(\textbf{y}_{t}|\textbf{x}_{t}^i; \theta) p(\textbf{x}_t^i|\tilde{\textbf{x}}_{t-1}^i, \textbf{a}_t; \theta)}{q(\textbf{x}_t^i|\tilde{\textbf{x}}_{t-1}^i, \textbf{y}_{t}, \textbf{a}_{t}; \theta)}$;
 			\STATE Normalize weights $w_{t}^i=\frac{w_{t}^i}{\sum_{m=1}^{N_p}w_{t}^m}$ for $i\in[N_p]$;
 			\STATE  Resample $\{\textbf{x}_{t}^i,w_{t}^i\}_{i=1}^{N_p}$ to obtain $\{\tilde{\textbf{x}}_{t}^i,\tfrac{1}{N_p}\}_{i=1}^{N_p}$;
 			\ENDFOR
 			\STATE Calculate the loss function $\mathcal{L}_{\text{overall}}$;
 			\STATE Update $\theta$ by gradient descent:
 				$\theta\leftarrow \theta-\gamma\nabla_\theta\mathcal{L}_{\text{overall}}$\,\,.\\
 			\ENDWHILE	
 		\end{algorithmic}
 	\end{algorithm}
 	
    We provide in Algorithm~\ref{alg:DPFs-CM} a DPF framework where the proposed measurement model is applied, which we name as differentiable particle filter with conditional normalizing flow measurement model (DPF-CM).
 	 	
	\section{Experiment Results}
	\label{sec:experiment_results}
	In this section, we compare the performance of the proposed measurement model with other measurement models used in several state-of-the-art works on DPFs~\cite{jonschkowski18,wen2021end,karkus2018particle} in a synthetic visual tracking task introduced in~\cite{kloss2021train,haarnoja2016backprop}. Two baselines are used in this experiment, including the vanilla differentiable particle filter (DPF) and the conditional normalizing flow differentiable particle filter (CNF-DPF)~\cite{chen2021differentiable}. We name the two baselines equipped with the proposed measurement model as the DPF-CM and the CNF-DPF-CM, respectively\footnote{Code to reproduce the experiment results is available at:~\url{https://github.com/xiongjiechen/Normalizing-Flows-DPFs}.}.

	\subsection{Experiment Setup}
\begin{figure}[h]
		\begin{center}
			\includegraphics[width=0.42\linewidth]{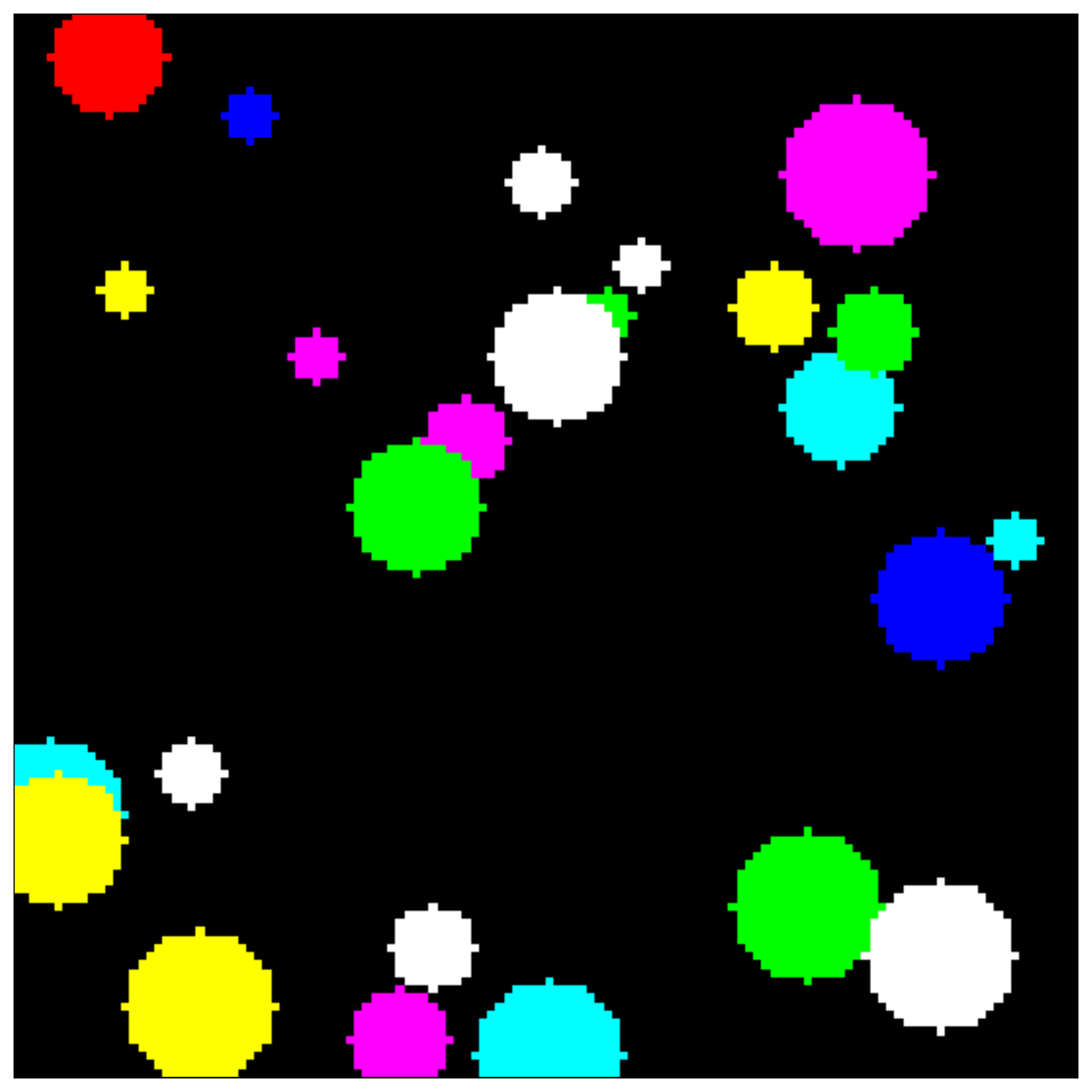}
			\caption{An observation image in the disk tracking experiment. The goal is to track the red disk among distractors.}
			\label{fig:red_disk}
		\end{center}
	\end{figure}

	The goal of this visual tracking problem is to track the location of a moving red disk. There are two key challenges in this experiment. Firstly, there are 25 distracting disks in observation images, and the distractors move as well while we try to locate the targeted red disk. Secondly, since collisions are not considered in this environment, the red disk may be occluded by distractors or move out of the boundary. The color of these distractors are randomly drawn from the set of \{green, blue, cyan, purple, yellow, white\}, and the radii of them are randomly sampled with replacement from \{3, 4, $\cdots$, 10\}. The radius of the tracking objective is set to be 7. Note that the locations of the disks, including the target and distractors, are initialized following a uniform distribution over the observation image. The initial velocity of the disks are sampled from a standard Gaussian distribution. Figure~\ref{fig:red_disk} shows an example of the observation image at time step $t=0$, which is an RGB image with the dimension 128$\times$128$\times$3.
	\begin{figure}[th]
		\begin{center}
			\includegraphics[width=\linewidth]{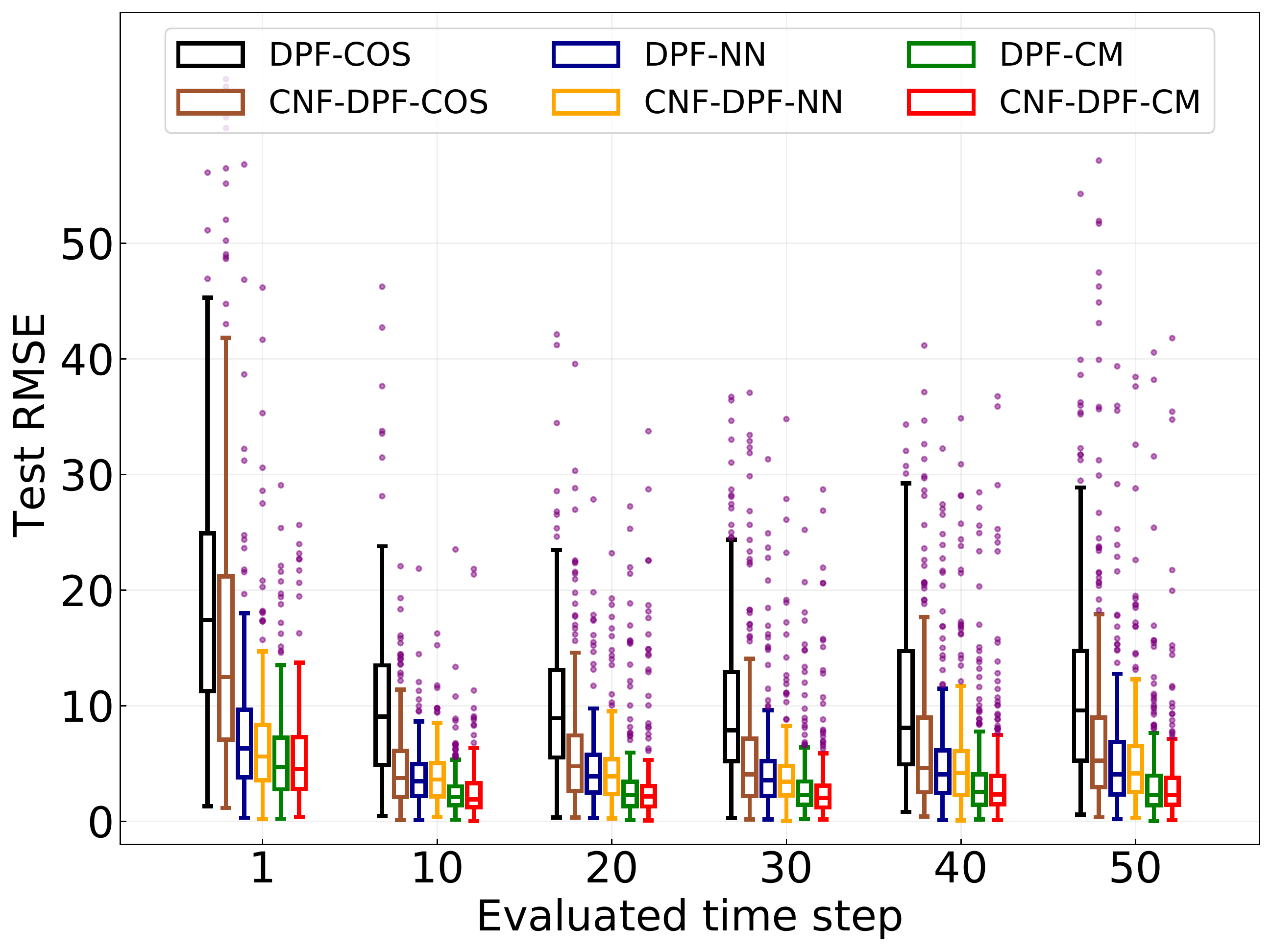}
			\caption{A boxplot that shows test RMSEs of different methods evaluated at specified time steps. The test set consists of 50 different trajectories, each with 50 time steps. Data points in the boxplot are generated from 5 simulation runs with different random seeds.}
			\label{fig:test_rmse_boxplot_all}
		\end{center}
	\end{figure}
	
	With a slight abuse of notations, in this example, we denote by $\textbf{x}_t$ the location of the red disk at the $t$-th time step, and $\textbf{a}_t$ the action. The velocity of the red disk at the $t$-th time step and the dynamic of the targeted red disk can be described as follows~\cite{kloss2021train}:
	\begin{align}
		\hat{\textbf{a}}_t &= \textbf{a}_t + \epsilon_t,\;\;\;\;\epsilon_t\overset{\text{i.i.d}}{\sim}\mathcal{N}(\textbf{0},\;\; \sigma_\epsilon^2\mathbb{I})\,\,,\\ 
		\textbf{x}_{t+1} &= \textbf{x}_t + \hat{\textbf{a}}_t + \alpha_t,\;\;\;\;\alpha_t\overset{\text{i.i.d}}{\sim}\mathcal{N}(\textbf{0},\;\; \sigma_\alpha^2\mathbb{I})\,\,,
		\label{eq:dynamic1}
	\end{align}
	where $\hat{\textbf{a}}_t$ is the noisy action obtained by adding random action noise $\epsilon_t$, $\sigma_\epsilon=4$ is the standard deviation of the action noise, and $\alpha_t$ is the dynamic noise whose standard deviation is $\sigma_\alpha=2$. The distractors follow the same dynamic presented above.

 We model the prior distribution $p_Z(\cdot)$ as a standard Gaussian distribution $\mathcal{N}(\textbf{0},\; \mathbb{I})$. We choose to use the conditional Real-NVP model~\cite{winkler2019learning} to construct conditional normalizing flow $\mathcal{T}_\theta(\cdot)$. An optimal transport map-based resampling approach~\cite{corenflos2021differentiable} is adopted for the resampling step in Line 13 of Algorithm~\ref{alg:DPFs-CM}. For the proposal distribution $q(\textbf{x}^i_t|\tilde{\textbf{x}}_{t-1}^i, \textbf{y}_t, \textbf{a}_t;\theta)$, we follow the setup in~\cite{chen2021differentiable} where the proposed particles are generated through conditional normalizing flows.

 We optimized the DPFs end-to-end by minimizing a loss function
 	$\mathcal{L}_{\text{overall}}$ which consists of $\mathcal{L}_{\text{RMSE}}$,
 	the root mean square error
 	(RMSE) between the prediction and the ground truth, and $\mathcal{L}_{\text{AE}}$, the autoencoder reconstruction loss 
 	w.r.t observations:
 	\begin{gather}
 		\label{eq:loss_overall}
 		\mathcal{L}_{\text{overall}}=\mathcal{L}_{\text{RMSE}}+\mathcal{L}_{\text{AE}}\,\,,\\
 		\label{eq:loss_rmse}
 		\mathcal{L}_{\text{RMSE}}=\sqrt{\frac{1}{T}\sum_{t=0}^{T}||\hat{\textbf{x}}_t-\textbf{x}_t^*||_2^2}\,\,,\\
 		\label{eq:loss_ae}
 		\mathcal{L}_{\text{AE}}=\frac{1}{T}\sum_{t=0}^{T}||D_\theta(E_\theta(\textbf{y}_t))-\textbf{y}_t||_2^2\,\,.
 	\end{gather}
 	In Eqs.~\eqref{eq:loss_overall},~\eqref{eq:loss_rmse}, and~\eqref{eq:loss_ae}, $T$ is the number of time steps, $||\cdot||_2$ denotes the $L^2$ norm, $x^*_t$ is the ground truth state at the $t$-th time step, $\hat{\textbf{x}}_t=\sum_{i=1}^{N_p}w_t^i\textbf{x}_t^i$ is an estimation of $\mathbb{E}_{p(\textbf{x}_t|\textbf{y}_{1:t})}[\textbf{x}_t]$, and $D_\theta(\cdot)$ and $E_\theta(\cdot)$ are respectively the decoder and the encoder of an autoencoder where the latent variable $\textbf{e}_t = E_\theta(\textbf{y}_t)$ is a 32-dimensional vector.

	\subsection{Experiment results}

\begin{figure}[t]
		\begin{center}
			\includegraphics[width=\linewidth]{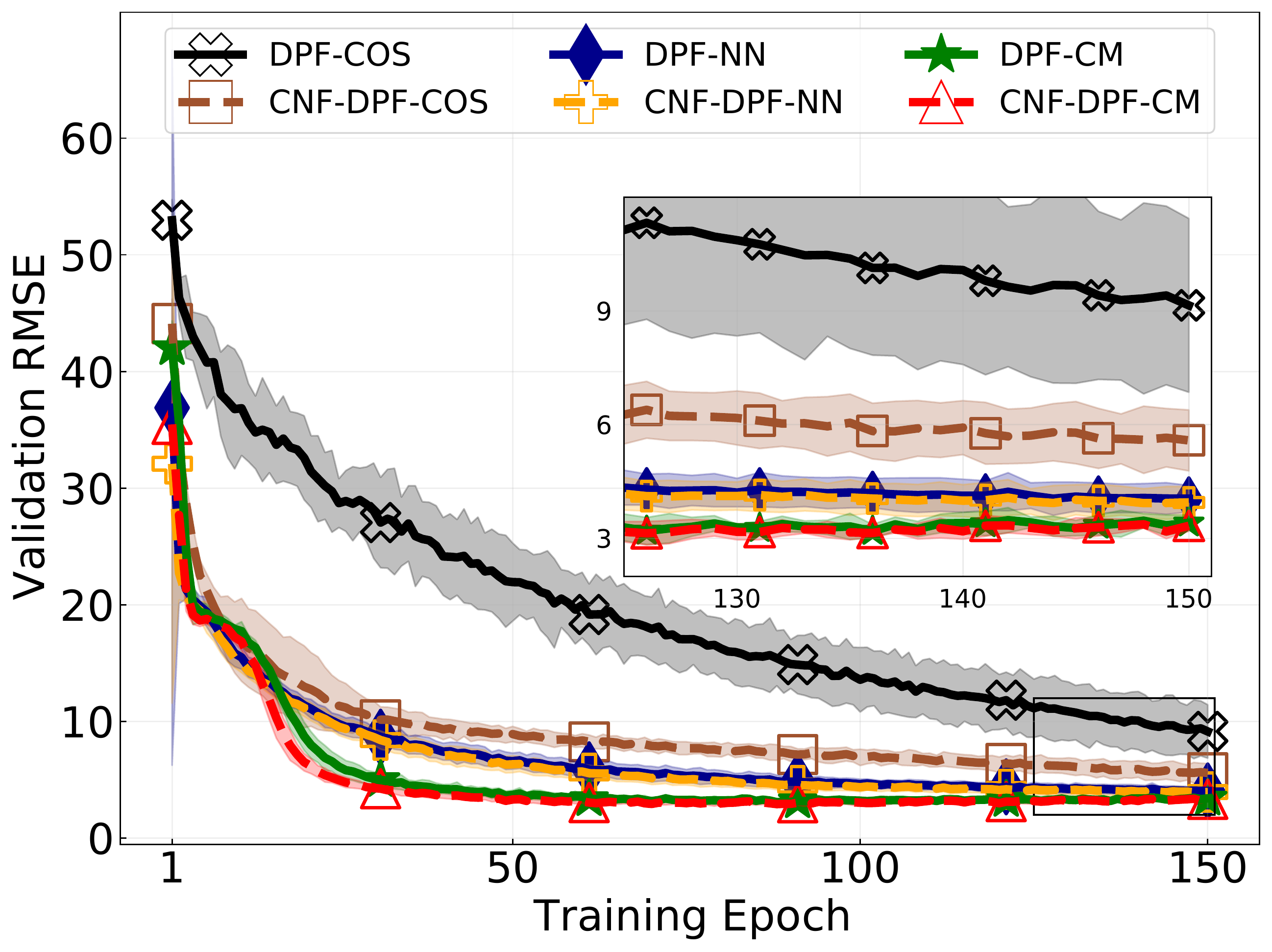}
			\caption{Validation RMSE of different approaches during the training process. The validation set consisting of 50 different trajectories of length 50. The error bar represents the 95\% confidence interval of validation RMSE among 5 simulation runs.}
			\label{fig:validation_loss_all}
		\end{center}
	\end{figure}

    In reporting and discussing the experiment results with Fig.~\ref{fig:test_rmse_boxplot_all}, Fig.~\ref{fig:validation_loss_all} and Table~\ref{tab:rmse}, we adopt the following naming convention to denote the proposed method and the compared algorithms -- ``DPF" refers to a baseline DPF~\cite{wen2021end}; the prefix ``CNF-" refers to the approach introduced in~\cite{chen2021differentiable} where conditional normalizing flows are used to construct dynamic models and proposal distributions; the suffix ``-COS" represents the measurement model proposed in~\cite{wen2021end} which considers the cosine similarity between encoded features as the likelihood; the suffix ``-NN" is used to denote the measurement model introduced in~\cite{karkus2018particle,jonschkowski18}, which outputs the likelihood via a neural network with encoded features as its input; ``-CM" denotes the proposed method.

	Fig.~\ref{fig:test_rmse_boxplot_all} compares RMSEs of state predictions from different methods on the test set.
	It can be observed that the DPFs with the proposed meausrement model, i.e. the DPF-CM and CNF-DPF-CM, exhibit the lowest RMSEs. The overall test RMSEs averaged across 50 time steps are reported in Table~\ref{tab:rmse}, which shows that both the DPF-CM and CNF-DPF-CM produce the smallest overall test RMSEs. In addition, Fig.~\ref{fig:validation_loss_all} plots the validation RMSEs recorded during the training phase. The proposed measurement model leads to faster convergence and the smallest validation RMSEs.

	\begin{table}[ht]
		\caption{The comparison of the overall average RMSE on the test set between different approaches, the standard deviation is calculated through 5 simulation runs.}
		\centering
		\begin{tabular}{c|c||c|c}
			\hline
			\textbf{Method} & \textbf{RMSE}&\textbf{Method} & \textbf{RMSE}  \\ \hline
			DPF-COS &  $10.61 \pm 2.22$&CNF-DPF-COS &  $5.87 \pm 0.83$ \\ \hline
			DPF-NN & $4.26\pm 0.32$&CNF-DPF-NN &  $4.19 \pm 0.31$ \\ \hline
			DPF-CM  & $\textbf{2.99}\pm \textbf{0.13}$&CNF-DPF-CM &  $\textbf{2.86} \pm \textbf{0.10}$ \\ \hline
		\end{tabular}
		\label{tab:rmse}
	\end{table}

	\section{Conclusion}
	\label{sec:conclusion}
	We propose in this paper a novel measurement model for differentiable particle filters. The proposed measurement model employs conditional normalizing flows to construct flexible and valid probability densities for likelihood estimation that can be embedded into existing DPF frameworks. A numerical implementation of the proposed method is provided in the paper, and we evaluate the performance of the proposed method in a visual disk tracking task. Numerical results show that the proposed method can significantly improve the tracking performance compared with state-of-the-art DPF variants.
	\bibliography{ref.bib} 
	\bibliographystyle{IEEEtran}
\end{document}